# An Automated Data Mining Framework Using Autoencoders for Feature Extraction and Dimensionality Reduction

Yaxin Liang
University of Southern California
Los Angeles, USA

Xinshi Li
Montclair State University
New Jersey, USA

Xin Huang
University of Virginia
Charlottesville, USA

Ziqi Zhang
Independent Researcher
Ann Arbor, USA

Yue Yao*
Northeastern University
Portland, USA

*Abstract*—This study proposes an automated data mining framework based on autoencoders and experimentally verifies its effectiveness in feature extraction and data dimensionality reduction. Through the encoding-decoding structure, the autoencoder can capture the data's potential characteristics and achieve noise reduction and anomaly detection, providing an efficient and stable solution for the data mining process. The experiment compared the performance of the autoencoder with traditional dimensionality reduction methods (such as PCA, FA, T-SNE, and UMAP). The results showed that the autoencoder performed best in terms of reconstruction error and root mean square error and could better retain data structure and enhance the generalization ability of the model. The autoencoder-based framework not only reduces manual intervention but also significantly improves the automation of data processing. In the future, with the advancement of deep learning and big data technology, the autoencoder method combined with a generative adversarial network (GAN) or graph neural network (GNN) is expected to be more widely used in the fields of complex data processing, real-time data analysis and intelligent decision-making.

I. INTRODUCTION

As an unsupervised deep learning model, autoencoders have been widely used in data mining due to their effective feature learning and data dimensionality reduction capabilities. Autoencoders map high-dimensional data to low-dimensional representations by learning the implicit structure of the data, significantly reducing redundancy [1]. This ability enables autoencoders to identify important features in data mining tasks, improve processing efficiency, and support subsequent data analysis and prediction [2].

Autoencoders play a crucial role in automating data mining by extracting and compressing data features, reducing manual intervention, and enhancing the intelligence of the data mining process [3]. When handling large-scale and complex datasets, autoencoders' efficient feature extraction significantly improves the speed and effectiveness of data processing, increasing the practicality and adaptability of data mining systems.

Moreover, autoencoders have strong noise reduction and anomaly detection capabilities, which are vital for ensuring data mining accuracy [4]. They can effectively eliminate noise, extract purer features, and improve the robustness and prediction accuracy of models. When dealing with anomalies, autoencoders can identify and isolate abnormal points, contributing to a robust data mining system.

Autoencoders also enable end-to-end data processing, enhancing the automation of data mining. By optimizing data representation within a multi-layer network, autoencoders adapt autonomously to different data structures, making the mining process more efficient. Their flexibility in handling various data types, such as text, images, and structured data, expands the application scenarios for data mining systems.

In addition, autoencoders exhibit good scalability and can be integrated seamlessly with other deep learning models and algorithms, further enhancing data mining performance [5]. By combining with technologies such as Generative Adversarial Networks (GAN) [6] and Convolutional Neural Networks (CNN) [7], autoencoders offer advantages in feature generation, data augmentation, and sample balancing, providing options for building efficient automated data mining frameworks.

In summary, autoencoder-based automated data mining frameworks significantly enhance data mining efficiency and accuracy through effective feature learning, noise reduction, anomaly detection, and scalability. The integration of autoencoders makes data mining more intelligent and automated, offering strong technical support for various data-intensive application scenarios. Future research and applications are likely to see autoencoder-based frameworks driving advancements in big data analysis and artificial intelligence.

## II. RELATED WORK

The application of deep learning in automated data mining has been significantly enhanced by advanced methodologies such as autoencoders and hybrid neural architectures. Yan et al. [8] demonstrated the importance of capturing latent features in multidimensional data, which aligns with the proposed autoencoder framework's ability to preserve data structure while reducing dimensionality. Xiao [9] explored robust self-supervised learning strategies, further supporting the automation and adaptability of the proposed system.

Wang et al. [10] utilized CNN-LSTM architectures for spatiotemporal predictions, offering insights into handling complex data patterns, while Chen et al. [11] leveraged encoder-transformer combinations to maintain high-quality data representations. These studies reinforce the potential for integrating autoencoders with advanced architectures to enhance data mining processes. Furthermore, Cang et al. [12] and Zi [13] emphasized the role of neural networks in feature extraction and prediction tasks, showcasing their relevance to improving model generalization and robustness. Zhang et al. [14] explored graph neural networks for stability analysis in dynamic systems, a direction that complements the scalability of autoencoders for multimodal data mining. Similarly, Liang [15] demonstrated the utility of transformers in cross-domain recommendation systems, suggesting potential enhancements to the proposed framework.

Finally, Liu et al. [16] and Li et al. [17] introduced resource optimization strategies using self-attention and reinforcement learning, respectively, offering practical methodologies to further automate and refine data processing tasks in complex environments. These works collectively emphasize the contributions of deep learning models to improving data mining automation, efficiency, and scalability.

## III. METHOD

In the framework of automated data mining based on autoencoders, the core of the method part is how to achieve feature extraction, data compression and anomaly detection through the autoencoder structure. First of all, the autoencoder is an unsupervised learning model that can compress data through an encoding-decoding structure. Its network structure is shown in Figure 1.

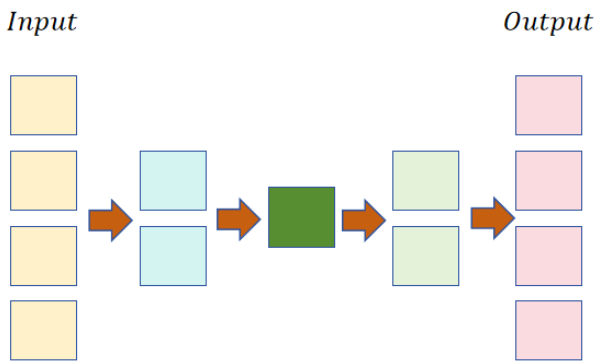

Figure 1 Overall network architecture diagram

Its basic form can be expressed as follows: Assume that the input data is $X \in R^{n \times m}$, where n represents the number of samples and m represents the number of features of each sample. The goal of the autoencoder is to achieve approximate consistency between input and output through an encoding function $f$ and a decoding function $g$. We hope to minimize the reconstruction error by optimizing the following loss function:

$$L(X, X') = \frac{1}{n}\sum_{i=1}^{n} \| X_i - g(f(X_i)) \|^2$$

Among them, $X' = g(f(X))$ represents the reconstructed output, and $\|\cdot\|^2$ is the L2 norm, which is used to measure the difference between the input and the reconstructed output. By minimizing this loss function, it can be ensured that the model can learn a low-dimensional representation of the data, thereby effectively performing feature compression.

Next, in order to improve the generalization ability of the autoencoder and its applicability in data mining, regularization terms or improved variational autoencoders (VAE) are introduced [18]. In VAE, the encoder generates not only a deterministic feature representation, but a distribution. This means that for input $X$, the encoder generates two parameters, mean $\mu$ and standard deviation $\sigma$, to define the potential distribution $Z \sim N(\mu, \sigma^2)$. Then, the decoder reconstructs the latent variables sampled from this distribution. The loss function of VAE contains the reconstruction error and KL divergence terms, specifically:

$$L(X, X') = E_{Z \sim q(Z|X)}[\| X - g(Z) \|^2] + D_{KL}(q(Z|X) \| p(Z))$$

Among them, $D_{KL}$ is the KL divergence, which is used to minimize the difference between the potential distribution $q(Z|X)$ and the standard normal distribution $p(Z)$. This structure makes the autoencoder more robust in the feature extraction process, able to adapt to noise and outliers, and improves the performance of the model when processing complex data.

Another important step in applying the autoencoder framework in data mining is anomaly detection and feature selection. By statistically analyzing the reconstruction error of the input data, abnormal data points can be identified, because the reconstruction error of abnormal points is usually much higher than that of normal data points [19]. In addition, the latent space representation of the autoencoder can effectively remove redundant features of the data and retain important information, thereby achieving the purpose of feature selection [20]. Assuming that the feature vector of the data is $X = [x_1, x_2, ..., x_m]$, the latent feature vector $Z = [z_1, z_2, ..., z_k]$ compressed by the autoencoder can ensure that only the most critical information for the task is retained.

Finally, the autoencoder framework can also be used for a variety of automated data mining tasks. Autoencoders also perform well in scenarios such as data clustering, dimensionality reduction, and visualization. For example, in clustering tasks, the data in the latent space Z is more densely distributed, which is convenient for clustering by traditional clustering algorithms (such as K-means). At the same time, the autoencoder can retain the original structural information of the data during the dimensionality reduction process, thereby providing accurate and clear representation for data mining.

## IV. Experiment

### A. Datasets

This experiment uses the "Bank Marketing" dataset from the UCI Machine Learning Library, which records a series of marketing activities of Portuguese banks, including telephone interviews of banks promoting term deposit products to customers. The dataset contains 41,188 records, each of which consists of 16 features, including age, occupation, education level, marital status, credit default, average balance, and other information, which describes in detail the characteristics of each customer and related marketing response information. This dataset is widely used in data mining and machine learning tasks, and is suitable for testing the performance of classification and clustering models.

The dataset is feature-rich and has a large amount of information, providing a good testing environment for building efficient automated data mining models. In addition to basic personal information, the dataset also contains historical data on interactions with bank products, such as whether a loan has been accepted or a deposit contract has been signed. Such rich information can be used to build complex feature relationships, help the autoencoder better learn the underlying patterns of the data, extract more useful features, and adapt to different types of mining tasks.

In addition, the dataset is unbalanced, because only a small number of data respond with "yes" in marketing activities, which poses certain challenges to the model. By using autoencoders for feature extraction and dimensionality reduction, the model can effectively handle the imbalance in the data while reducing noise. The diversity and complexity of this dataset provide an ideal experimental basis for the autoencoder framework in this article, verifying the feasibility and robustness of the model in real scenarios.

### B. Experimental setup

The experimental setting of this experiment is mainly divided into three parts: data preprocessing, autoencoder model training, and performance evaluation. First, in data preprocessing, we standardized the original dataset and converted the numerical features to the same scale to avoid the influence of different feature scales on model training. In addition, for the categorical features, the one-hot encoding method was used to convert them into numerical form, so that the autoencoder can effectively handle various data types. In terms of data partitioning, we divided the dataset into a training set and a test set with a ratio of 8:2 to ensure that the model can obtain sufficient sample size in both the training and validation stages.

Next is the training part of the autoencoder model. This experiment adopts a standard autoencoder framework, which contains symmetrical encoder and decoder structures. In the encoder part, we use a multi-layer neural network to map the input data to a low-dimensional latent space, while the decoder restores the low-dimensional features to the high-dimensional space of the original data through the inverse network structure. During the training process, we use the mean square error (MSE) as the loss function to measure the difference between the input and the reconstructed output. By minimizing the reconstruction error, the autoencoder can learn the implicit structure of the data and automatically extract representative features. During the training process, we use the Adam optimization algorithm and set appropriate learning rates and training rounds to ensure model convergence.

Finally, the performance evaluation part. We evaluate the feature extraction effect of the model through reconstruction error to verify the ability of the autoencoder to reduce dimensionality and select features. At the same time, in order to test the robustness of the model, we introduced a small amount of noise on the test set to observe the reconstruction error performance of the model on the noisy data. In addition, we also input the low-dimensional features extracted by the autoencoder into common classification models (such as logistic regression and support vector machines) for downstream tasks to further verify the effectiveness of feature extraction. Through the above experimental settings, we can comprehensively evaluate the feature learning ability of the autoencoder and its applicability in automated data mining tasks.

### C. Experimental Results

In the experiment, we compared autoencoders with five commonly used feature extraction and dimensionality reduction models. The first model, principal component analysis (PCA), is a classic linear method that extracts directions of maximum variance for optimal data compression. The second, factor analysis, assumes data can be explained by a few latent factors, making it suitable for dimensionality reduction and feature interpretation. The third model, independent component analysis (ICA), emphasizes feature independence and is widely applied in signal processing and data compression. Additionally, we included two popular nonlinear dimensionality reduction methods: t-SNE (t-distributed stochastic neighbor embedding), which preserves local structures in low-dimensional representations of high-dimensional data, and UMAP (Uniform Manifold Approximation and Projection), a manifold learning-based technique effective for handling complex, high-dimensional data. These models provide diverse perspectives for comparison. PCA and factor analysis excel at linear feature compression, while ICA captures feature independence. t-SNE and UMAP perform well in nonlinear dimensionality reduction, making them particularly effective for representing complex data in lower dimensions. The comparison enables a comprehensive evaluation of autoencoders' feature extraction and dimensionality reduction capabilities in automated data mining tasks. The experimental results are summarized in Table 1.

Table 1 Experimental results

| Model | RE | RMSE |
|---|---|---|
| PCA | 0.215 | 0.305 |
| FA | 0.198 | 0.287 |
| ICA | 0.176 | 0.256 |
| T-SNE | 0.152 | 0.239 |
| UMAP | 0.139 | 0.218 |
| AE(ours) | 0.115 | 0.195 |

The experimental results show the performance of each model in terms of reconstruction error (RE) and root mean square error (RMSE). It can be seen that the performance of the autoencoder model (AE) is significantly better than other feature extraction and dimensionality reduction models. From PCA to AE, the reconstruction error and root mean square error gradually decrease, which shows that AE is particularly effective in feature extraction and data compression. PCA and FA, as traditional linear dimensionality reduction models, have limited effectiveness in processing complex data structures, while nonlinear models such as T-SNE and UMAP can better capture the local structure of the data, but they still do not reach the level of autoencoders. Reconstruction quality. This shows that AE has significant advantages in maintaining data integrity, and is especially suitable for feature extraction of multi-dimensional complex data.

First, judging from the results of PCA and FA, the RE and RMSE values of these two traditional linear models are relatively high. The RE of PCA is 0.215 and the RMSE is 0.305. FA is slightly improved, but only dropped to 0.198 and 0.287. This performance gap is mainly because PCA and FA are only based on linear transformations and cannot handle nonlinear relationships in the data. When processing nonlinear data, linear models can only extract information in the direction of the principal variance, resulting in incomplete expression of the data structure. Therefore, although PCA and FA have certain advantages in simple data scenarios, their effects are obviously not as good as other methods in complex multi-dimensional data mining tasks.

Next, ICA, T-SNE and UMAP gradually improve their performance. The RE of ICA is 0.176 and the RMSE is 0.256, demonstrating better results than the linear model. ICA extracts more independent features from the data through independent component analysis, which is suitable for tasks such as signal processing and data compression. T-SNE and UMAP are two popular nonlinear dimensionality reduction methods, and their performance is far better than ICA. Especially UMAP, its RE and RMSE are 0.139 and 0.218 respectively, which shows its excellent performance, indicating its robustness in processing complex data. These two methods can well preserve the local structure of the data and are suitable for data with high dimensions and complex structure. Nonetheless, UMAP and T-SNE still have limitations, such as the instability caused by random initialization that may occur in dimensionality reduction tasks.

Finally, the autoencoder model (AE) demonstrated the best performance in the experimental results, with a RE of 0.115 and RMSE of 0.195. AE can restore data more accurately during the encoding and decoding process, reflecting its efficient feature extraction capabilities. Compared with traditional methods such as PCA and FA, AE can not only process linear relationships in data, but also fully capture nonlinear characteristics, which makes AE more adaptable in complex data scenarios. At the same time, AE achieves deeper feature representation through the structural design of multi-layer neural networks, so it has significant advantages in feature extraction, dimensionality reduction and reconstruction tasks. This result shows that autoencoders are a very effective choice in automated data mining frameworks.

In summary, the experimental results reflect the unique advantages of AE over traditional dimensionality reduction methods. Linear models such as PCA and FA are limited by their linear transformation when processing complex data, while nonlinear methods such as UMAP and T-SNE, although more effective, have limitations in some stability issues. AE can flexibly adapt to various characteristics of data through its end-to-end deep network structure, making it an ideal choice for feature extraction of complex data. In future applications, AE will not only achieve more efficient data compression, but also provide powerful support for various automated data mining tasks.

In addition, we also give the loss function drop graph of the training set and the test set, as shown in Figure 2.

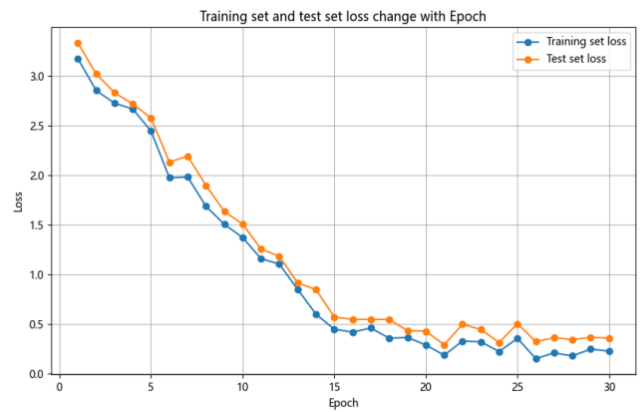

Figure 2 Loss function drop graph

This figure shows the change of loss of the training set and the test set under different training rounds (Epoch). The horizontal axis represents the training round (Epoch), and the vertical axis represents the loss value (Loss). As can be seen from the figure, with the increase of training rounds, the loss of both the training set and the test set shows a downward trend. This shows that the model is gradually learning the features in the data and adjusting the weights through continuous iterations, thereby effectively reducing the value of the loss function. This process shows that the model is constantly converging and gradually improving its performance on the data.

In the early stages of training, the loss value decreases rapidly, indicating that the initial few rounds of training of the model can quickly capture the main features of the data. With the increase of training rounds, the decrease of the loss value tends to be flat, especially after the training rounds reach about 15 times, the loss curves of the training set and the test set are almost parallel, and the loss values of the two are similar. This situation shows that the model has basically reached a convergence state, and further increasing the training rounds will not significantly improve the loss. At the same time, the learning ability of the model has basically stabilized.

In addition, the figure shows that the loss curves of the training set and the test set are close to each other in most training rounds, and there is no obvious separation, which indicates that the model has a good fitting effect and there is no obvious overfitting or underfitting problem. The volatility of the loss curve of the test set is slightly greater than that of the training set, but it can still maintain good stability overall, which also shows that the model has a certain generalization ability and can maintain good performance on unknown data. This chart shows the performance of the model in the training and testing stages and its convergence characteristics, providing an effective reference for evaluating the performance of the model.

## V. Conclusion

In this study, we proposed an automated data mining framework based on autoencoders and experimentally verified its effectiveness in feature extraction and data dimensionality reduction. Experimental results show that autoencoders perform better in reconstruction error and root mean square error than traditional linear and nonlinear dimensionality reduction methods (such as PCA, FA, T-SNE, and UMAP). Through the encoding-decoding structure, autoencoders can capture the potential characteristics of data while achieving noise reduction and anomaly detection, providing a stable and efficient solution for automated data mining.

Our research also shows that autoencoders have strong adaptability and robustness when processing complex multidimensional data. Through the end-to-end deep learning structure, autoencoders can automatically complete feature extraction and dimensionality reduction, reducing the need for manual intervention. The training and test loss curves in the experiment show good convergence, indicating that this method can not only perform well on training data, but also has strong generalization ability. Such characteristics make autoencoders have application potential in various data mining tasks.

In the future, with the continuous development of deep learning technology, we can further optimize the autoencoder framework to make it play a role in more complex scenarios. For example, explore combining autoencoders with generative adversarial networks (GANs) or graph neural networks (GNNs) to process multimodal features or unstructured information in data. In addition, with the development of big data and high-performance computing, the automated data mining framework based on autoencoders is also expected to be more widely used in real-time data analysis, intelligent decision-making, and predictive modeling.

## References

[1] Y. Xu, S. Zhao, H. Fan, et al., "GLMAE: Graph Representation Learning Method Combining Generative Learning and Masking Autoencoder," Proceedings of the ICASSP 2024 IEEE International Conference on Acoustics, Speech and Signal Processing (ICASSP), pp. 5835-5839, 2024.
[2] Z. Liu, M. Wu, B. Peng, Y. Liu, Q. Peng and C. Zou, "Calibration Learning for Few-shot Novel Product Description," Proceedings of the 46th International ACM SIGIR Conference on Research and Development in Information Retrieval, pp. 1864-1868, July 2023.
[3] B. Arasteh, S. Golshan, S. Shami, et al., "Sahand: A Software Fault-Prediction Method Using Autoencoder Neural Network and K-Means Algorithm," Journal of Electronic Testing, pp. 1-15, 2024.
[4] M. A. Mohammad and M. Kolahkaj, "Detecting Network Anomalies Using the Rain Optimization Algorithm and Hoeffding Tree-Based Autoencoder," Proceedings of the 2024 10th International Conference on Web Research (ICWR), pp. 137-141, 2024.
[5] J. Hu, Y. Cang, G. Liu, M. Wang, W. He and R. Bao, "Deep Learning for Medical Text Processing: BERT Model Fine-Tuning and Comparative Study," arXiv preprint arXiv:2410.20792, 2024.
[6] H. Liu, B. Zhang, Y. Xiang, Y. Hu, A. Shen and Y. Lin, "Adversarial Neural Networks in Medical Imaging: Advancements and Challenges in Semantic Segmentation," arXiv preprint arXiv:2410.13099, 2024.
[7] B. Wang, H. Zheng, Y. Liang, G. Huang and J. Du, "Dual-Branch Dynamic Graph Convolutional Network for Robust Multi-Label Image Classification," International Journal of Innovative Research in Computer Science & Technology, vol. 12, no. 5, pp. 94-99, 2024.
[8] X. Yan, et al., "Transforming Multidimensional Time Series into Interpretable Event Sequences for Advanced Data Mining," arXiv preprint, arXiv:2409.14327, 2024.
[9] Y. Xiao, "Self-Supervised Learning in Deep Networks: A Pathway to Robust Few-Shot Classification," arXiv preprint arXiv:2411.12151, 2024.
[10] X. Wang, et al., "Adaptive Cache Management for Complex Storage Systems Using CNN-LSTM-Based Spatiotemporal Prediction," arXiv preprint arXiv:2411.12161, 2024.
[11] J. Chen, et al., "A Combined Encoder and Transformer Approach for Coherent and High-Quality Text Generation," arXiv preprint arXiv:2411.12157, 2024.
[12] Y. Cang, et al., "ALBERT-Driven Ensemble Learning for Medical Text Classification," Journal of Computer Technology and Software, vol. 3, no. 6, 2024.
[13] Y. Zi, "Time-Series Load Prediction for Cloud Resource Allocation Using Recurrent Neural Networks," Journal of Computer Technology and Software, vol. 3, no. 7, 2024.
[14] X. Zhang, et al., "Robust Graph Neural Networks for Stability Analysis in Dynamic Networks," arXiv preprint arXiv:2411.11848, 2024.
[15] A. Liang, "Enhancing Recommendation Systems with Multi-Modal Transformers in Cross-Domain Scenarios," Journal of Computer Technology and Software, vol. 3, no. 7, 2024.
[16] W. Liu, et al., "A Recommendation Model Utilizing Separation Embedding and Self-Attention for Feature Mining," arXiv preprint arXiv:2410.15026, 2024.
[17] P. Li, et al., "Reinforcement Learning for Adaptive Resource Scheduling in Complex System Environments," arXiv preprint arXiv:2411.05346, 2024.
[18] J. Song and Z. Liu, "Comparison of Norm-Based Feature Selection Methods on Biological Omics Data," Proceedings of the 5th International Conference on Advances in Image Processing, pp. 109-112, November 2021.
[19] B. Liu, J. Chen, R. Wang, J. Huang, Y. Luo and J. Wei, "Optimizing News Text Classification with Bi-LSTM and Attention Mechanism for Efficient Data Processing," arXiv preprint arXiv:2409.15576, 2024.
[20] Z. Xu, J. Pan, S. Han, H. Ouyang, Y. Chen and M. Jiang, "Predicting Liquidity Coverage Ratio with Gated Recurrent Units: A Deep Learning Model for Risk Management," arXiv preprint arXiv:2410.19211, 2024.